\documentclass{article}

\usepackage{spconf,amsmath,epsfig,amsfonts}
\usepackage{caption}
\usepackage{subcaption}

\title{Self-Supervised Temporal Analysis of Spatiotemporal Data}

\name{Yi Cao, Swetava Ganguli, Vipul Pandey}
\address{
    Apple\\
    {\tt\small \{ycao4,swetava,vipul\}@apple.com}
}

\begin{document}

\maketitle

\begin{abstract}
There exists a correlation between geospatial activity temporal patterns and type of land use. A novel self-supervised approach is proposed to stratify landscape based on mobility activity time series. First, the time series signal is transformed to the frequency domain and then compressed into task-agnostic \textit{temporal embeddings} by a contractive autoencoder, which preserves cyclic temporal patterns observed in time series. The pixel-wise embeddings are converted to image-like channels that can be used for task-based, multimodal modeling of downstream geospatial tasks using deep semantic segmentation. Experiments show that temporal embeddings are semantically meaningful representations of time series data and are effective across different tasks such as classifying residential and commercial areas.
\end{abstract}

\section{Introduction}\label{introduction}
Data mining mobility data is a cost-efficient and scalable solution for geographic landscape survey\cite{8898836,ahas2010using}. But positional errors of GPS can easily propagate into the aggregates of GPS crumbs that are typical input into downstream models, resulting in false positives of landscape classification. Time series methods such as ARIMA and GARCH \cite{shumway2000time} are more robust approaches that can extract useful information while tolerating inherent noise in mobility data.

We use Discrete Fourier Transform (DFT) to expose cyclic components in activity time series and find that there is a correlation between type of land use and activity's cadence. This finding motivates us to further develop a framework that connects the duct between time series analysis and Convolutional Neural Network (CNN). Because time series analysis typically casts problems as one dimensional sequence that spans along time axis but CNN models spatial relationship between pixels (of surface dimensions). We propose a framework that encodes frequency vector into a vector of controllable size,  called temporal embeddings, that can be fed into CNN for segmentation tasks, which is able to capture the  dynamics of activity in time horizon while correlating neighboring activities in spatial dimensions, unlocking the potential of spatiotemporal dataset.  

The remaining of this paper is structured as follows. In Section \ref{signal},  activity footprint and a few representative temporal patterns are shown, followed by their DFT representations. A few characteristic frequencies are visualized to show the correlation between temporal patterns and land use.  In Section \ref{embedding}, a embedding framework is proposed that can be used by a segmentation framework to enable supervised learning. In Section \ref{applications}, we presents three use cases that use temporal pattern to classify landscape.

\section{Methodology} \label{signal}

\subsection{Temporal pattern characterization}
The mobility data contains GPS traces shared by users, which are sequence of locations with timestamp. A general analysis scheme is aggregating activities within \emph{tiles} resulting from WGS 84 projection\cite{epsg}. A target regions is modeled as a blanket of tiles of chosen granularity. Let \(t\) denote time, and \(\Delta t\) be aggregation interval, and \(T\) be number of time bins within observation time horizon. \(x_t\) represents the count of activities at time \(t\) of a tile. Aggregating all traces that pass through this tile leaves us a time series \([x_{0}, x_{1}, ..., x_{T-1}]\).

\begin{figure}
    \centering
    \includegraphics[width=0.43\textwidth]{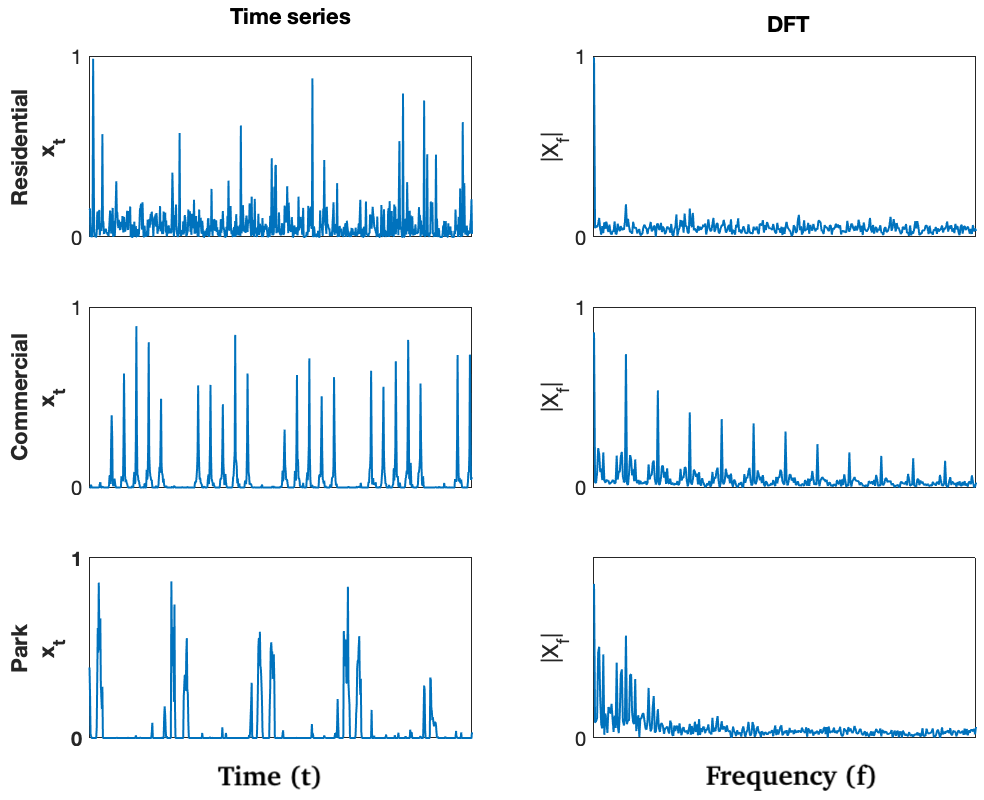}
     \caption{Normalized mobility data time series (left) and corresponding normalized DFT (right) in different land use.}
     \label{fig:probe_timeseries}
\end{figure}

It is clear that the shape of time series is correlated to land use based on the left column of Figure \ref{fig:probe_timeseries}. In residential area, activities appear to be uncoordinated therefore random. In contrast, time series from office present a much regular cadence. An inverse weekly pattern is observed in community park, which is in line to the fact that recreational facilities are usually busier on the weekend.

A typical time series analysis scheme is Discrete Fourier Transformation (DFT), which produces a sequence of complex numbers \([X_0, X_1, ..., X_{T-1} ]\), where:
\begin{equation}
X_k = \sum_{t=0}^{T-1}x_{t}e^{-\frac{i2\pi}{T}kt} \quad \quad k \in [0, T-1]
\end{equation}
DFT breaks down a signal with finite length into an array of sinusoidal signal of harmonic frequencies, with the fundamental frequency being \(f_0=\frac{1}{T\times \Delta t}\) and the rest being multiple of the fundamental frequency. The amplitude \(|X_{k}|\) measures significance of frequency \(f_k=\frac{k}{T\times \Delta t}\) among all harmonic frequencies. 

Based on the right column of Figure \ref{fig:probe_timeseries}, it is evident that DFT is still a noise-like signal when the input signal shows no clear pattern while cyclic signal results in spikes at positions of characteristic frequencies. It is worth noting that the first component of DFT is the sum of the input time series, therefore DFT encodes not only cadence information but also volume information. Any model that makes predictions based on volume can be enhanced by using DFT as input.

\setlength{\belowcaptionskip}{2pt}
\setlength{\textfloatsep}{2pt}
\begin{figure}
    \centering
    \begin{subfigure}{0.16\textwidth}
        \centering
	\includegraphics[width=\textwidth]{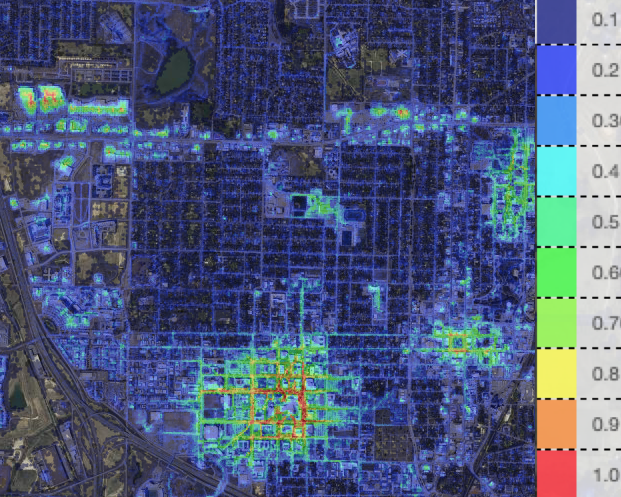}
	\caption{Daily pattern}
	\label{apple_daily}
    \end{subfigure}
    \hfill
    \begin{subfigure}{0.14\textwidth}
        \centering
	\includegraphics[width=\textwidth]{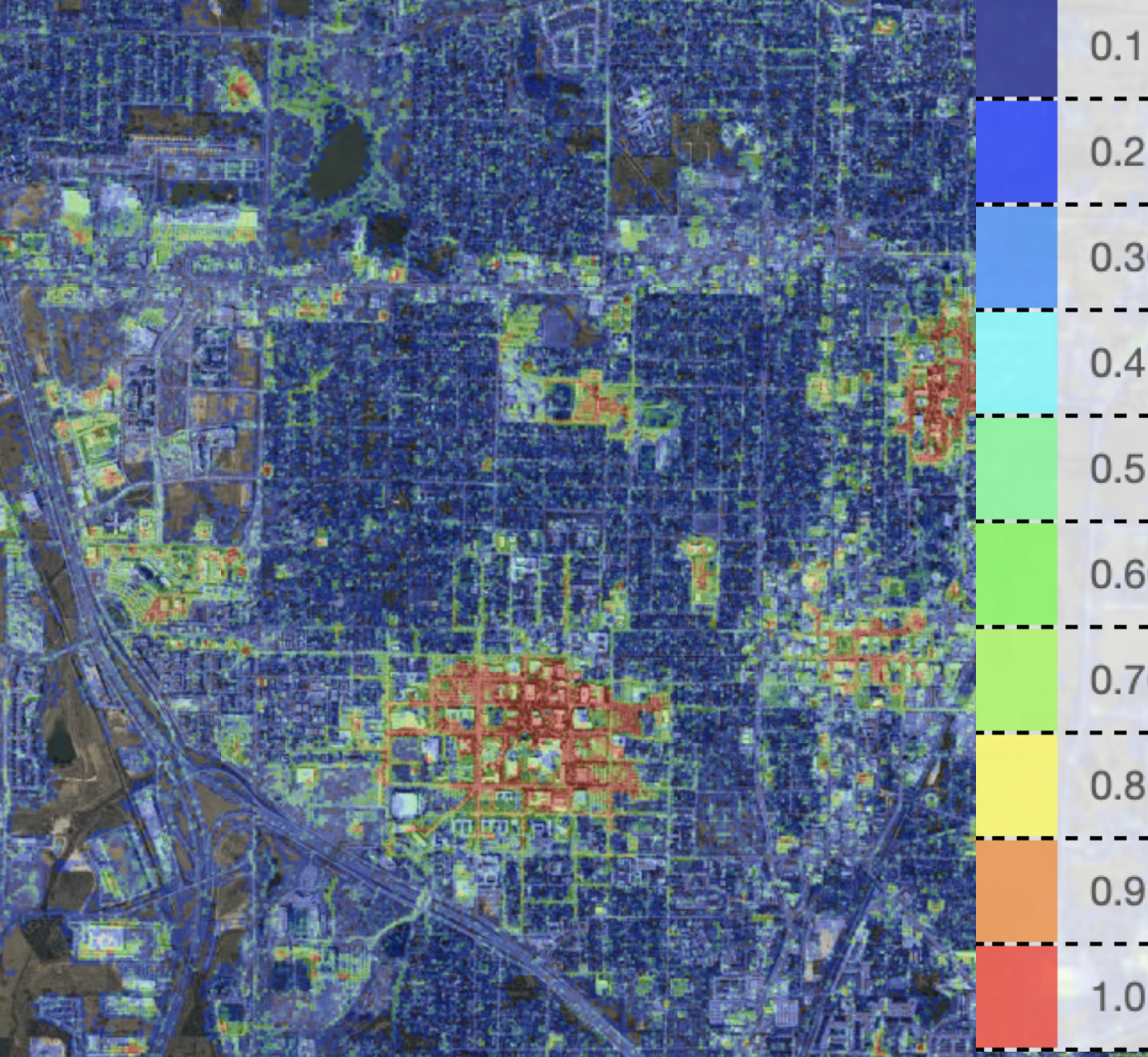}
	\caption{Weekly pattern}
	\label{apple_weekly}
    \end{subfigure}
    \hfill
    \begin{subfigure}{0.16\textwidth}
        \centering
	\includegraphics[width=\textwidth]{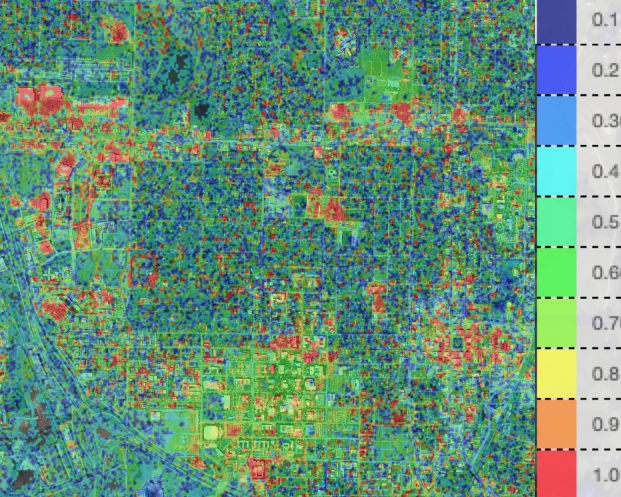}
	\caption{Yearly pattern}
	\label{apple_yearly}
    \end{subfigure}
    \caption{Heat map of DFT characteristic components.}
    \label{dft_comparison}
\end{figure}

Cadence of activities is an important temporal feature that characterizes geographic locations. If we properly adjust the value of $\Delta t$ and $T$, we could have the time series capture both short-term dynamics and long-term trending. For example, the DFT values can suggest if activities are recurrent on a daily, weekly, and even seasonal basis. Figure \ref{dft_comparison} shows geographic representation of a few characteristic frequencies where normalized DFT values are color-coded. The heat represents strong recurrent patterns. Some places have one or more recurrent patterns while others have none. Aside from these common temporal patterns, there are many other characteristic frequencies that are not necessarily perceivable to human but essential to special temporal patterns. 

We choose zoom 24 tiles ($1.5m\times1.5m$) for activity aggregation, each zoom 24 tile serves as a pixel characterized by a $d_{DFT}$-dimensional DFT vector. A contiguous, two-dimensional block of $256 \times 256$ zoom 24 tiles form a zoom 16 tile. With the DFT vector, a zoom-16 tile is presented as a $256 \times 256 \times d_{DFT}$ dimensional, image-like tensor that encodes the temporal activity pattern at that location. The tensor is a natural input format to deep learning architecture like CNN for classification or segmentation.

\subsection{Self-Supervised Temporal Representation Learning}\label{embedding}
\begin{figure*}
    \centering
    \includegraphics[width=\textwidth]{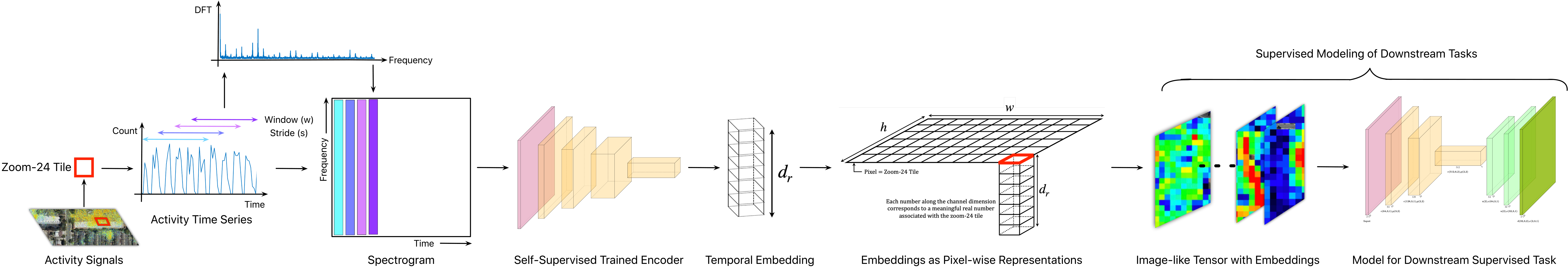}
    \caption{A schematic of the process for generating the temporal embeddings of zoom 24 tiles.}
    \label{fig::temporal_embeddings_schematic}
\end{figure*}

There are a few practical computational concerns using the raw DFT as feature. First, the size of the tensor will increase linearly with respect to $\Delta t$ and $T$. $d_{DFT}$ could be at the scale of tens of thousands if yearlong hourly time series are used, resulting in a tensor of tens of millions in size. Second, a DFT vector could be sparse or filled with small fractions typically at high frequencies which largely encode randomness and transient changes that do not help distinguish activity patterns in most of the cases.  Therefore, using raw DFT as input channel is not computationally efficient. Inspired by \cite{ganguli2022reachability}, we propose to employ an autoencoder to compress the $d_{DFT}$-dimensional DFT to derive more compact feature vectors of $d_r$ in size with minimal information loss, where $d_r << d_{DFT}$. The resulting artifact, called a \textit{temporal embedding}, is a task-agnostic representation analogous to spectral imaging for remote sensing, where channels corresponding to different spectral bands capture temporal information of a geographic location. 

Figure \ref{fig::temporal_embeddings_schematic} shows a schematic of the process of generating temporal embeddings for each zoom-24 tile and feeding them as pixel representations to CNN. The autoencoder architecture can be multi-layer perceptron-based, 1D convolution based, recurrent neural network-based, or transformer based, among other choices. In this paper, we restrict to using computer vision techniques for the self-supervised pretext task. The time series is converted to a spectrogram with an appropriate choice of window function, window length ($W$), and stride ($s$) between windows. Given a time series of length $T$, the length of the DFT for a window of size $W$ is $(W+1)/2$. The number of windows with stride, $s$, is $(T-W)/s+1$. In order to produce a square image, appropriate values can be calculated for $W$ and $s$ so that $(W+1)/2 = (T-W)/s+1$. A fully convolutional contractive autoencoder is trained to reconstruct the spectrogram for a time series and learns a compressed representation of the spectrogram in a lower-dimensional manifold. We call this manifold the temporal embedding field. Let $E(\cdot)$ and $D(\cdot)$ represent the encoder and decoder components of the autoencoder. Let $\Psi(z)$ be the spectrogram for the zoom-24 tile, $z$. The autoencoder is trained by minimizing the objective
\begin{equation}
    \mathcal{L}_{rec} = \|\Psi(z)-D(E(\Psi(z)))\|_2^2 
\end{equation}
\begin{equation}
    \mathcal{L}_{con} = \sum_{i=0}^{d_r-1} \|\nabla_{\Psi(z)} E(\Psi(z))\|_2^2 
\end{equation}
\begin{equation}
    \mathcal{L}_{tot} = \mathbb{E}_{z} \left[ \mathcal{L}_{rec} + \lambda \mathcal{L}_{con} \right].
\end{equation}

The optimal value of $\lambda=0.5$. The encoder of the trained autoencoder is then used to generate the compressed representation for each zoom-24 tile. The temporal embedding for a zoom-24 tile is computed as $E(\Psi(z))$. Increasing $d_r$ allows the contractive autoencoder to retain more semantically meaningful information that may be helpful for the reconstruction task. However,  larger $d_r$ results in lowering the maximum possible batch-size (governed by GPU memory) for training unimodal or multimodal models. To accommodate downstream classification task, We set $d_r=32$ given the practical constraints of heterogeneous compute clusters having GPUs with varied memory. 

\begin{figure}[h]
    \centering
    \begin{subfigure}{0.40\textwidth}
        \centering
        \includegraphics[width=\textwidth]{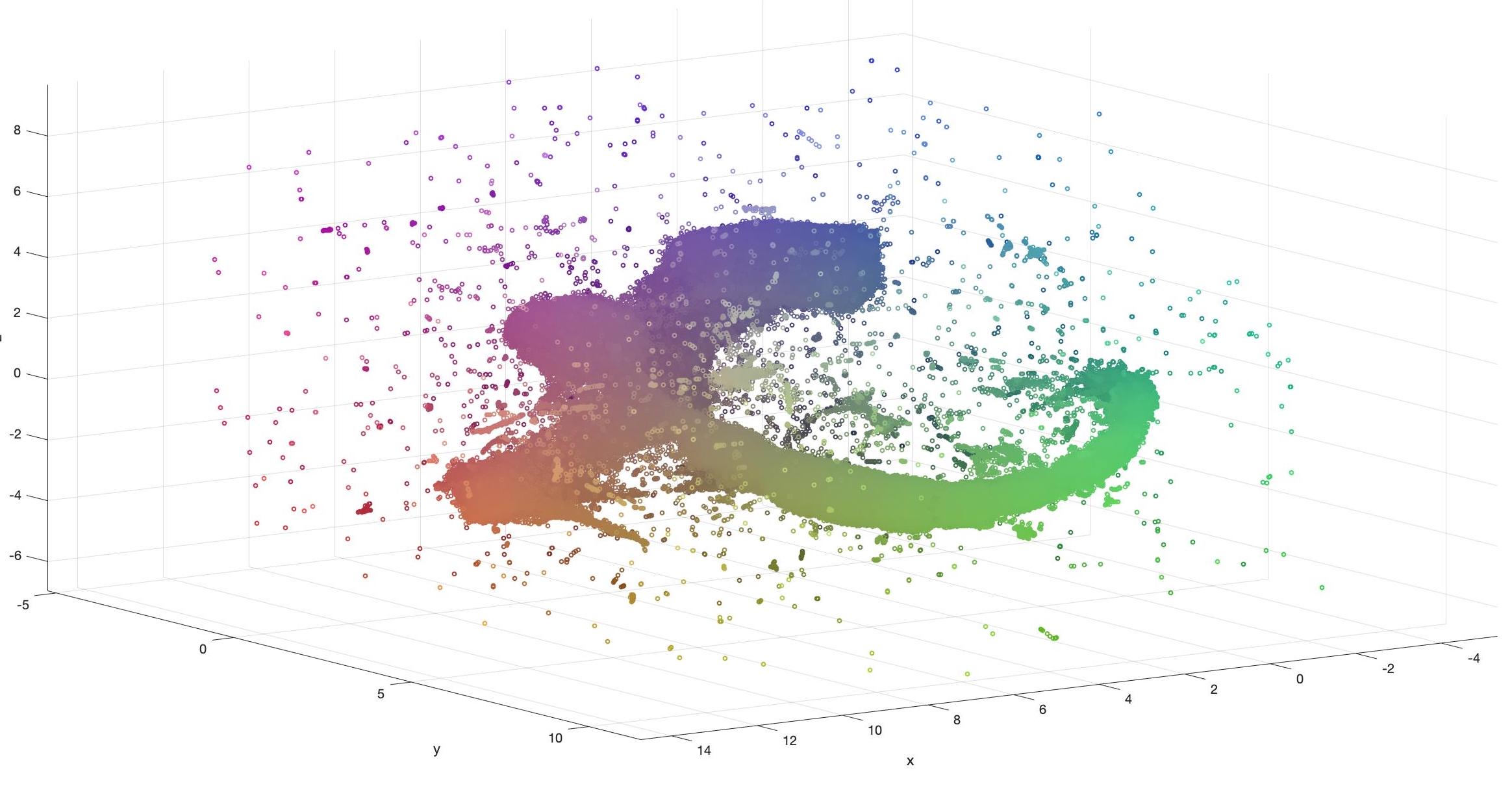}
        \caption{UMAP projection of learned temporal embeddings.}
        \label{umap}
    \end{subfigure}

    \begin{subfigure}{0.40\textwidth}
        \centering
        \includegraphics[width=\textwidth]{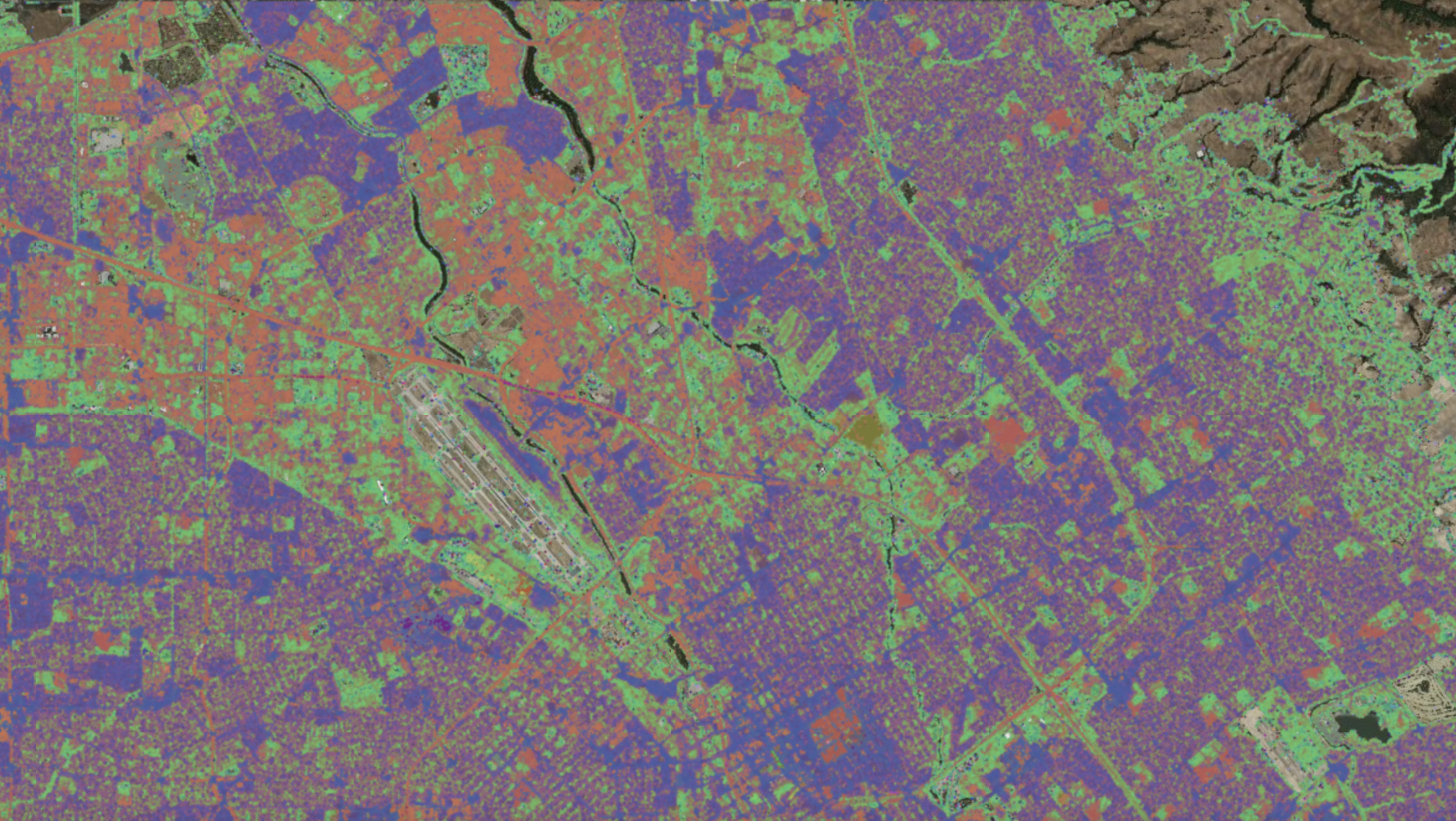}
        \caption{Temporal stratification of a geographic location.}
        \label{umap_viz}
    \end{subfigure}
    
     \caption{Landscape stratification using temporal embeddings.}
     \label{temporal_similarity}
\end{figure}

\begin{figure}[h]
    \centering
    \begin{minipage}{0.48\textwidth}
        \centering
        \includegraphics[width=0.225\textwidth]{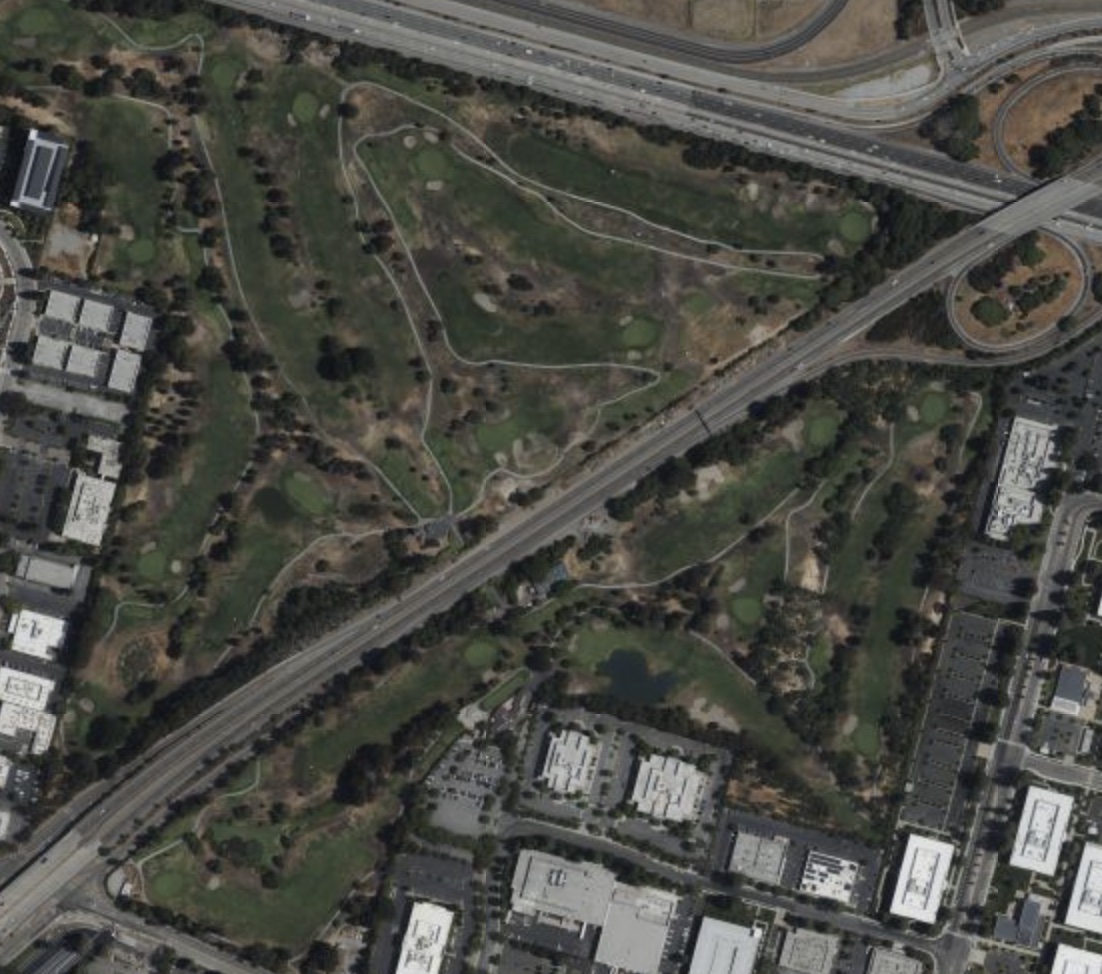}
        \includegraphics[width=0.225\textwidth]{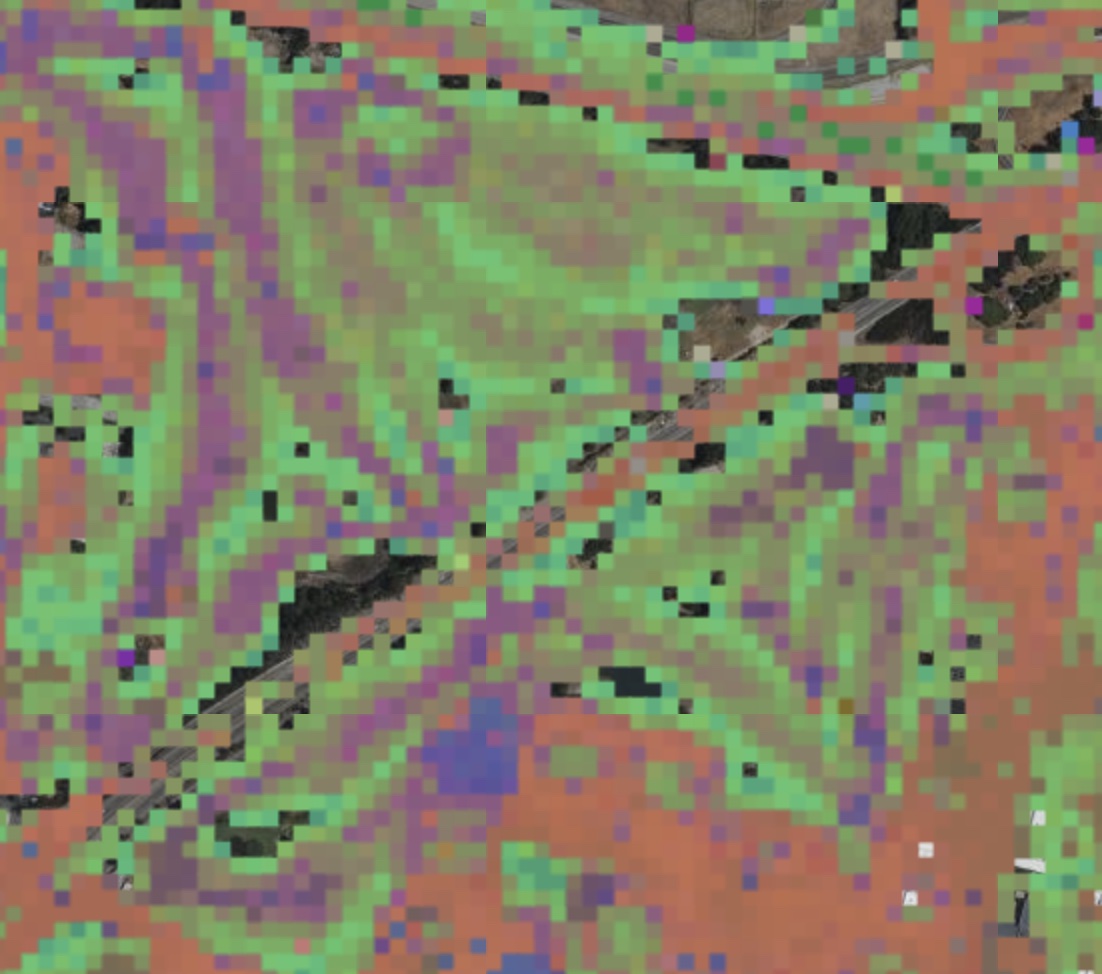}
        \includegraphics[width=0.225\textwidth]{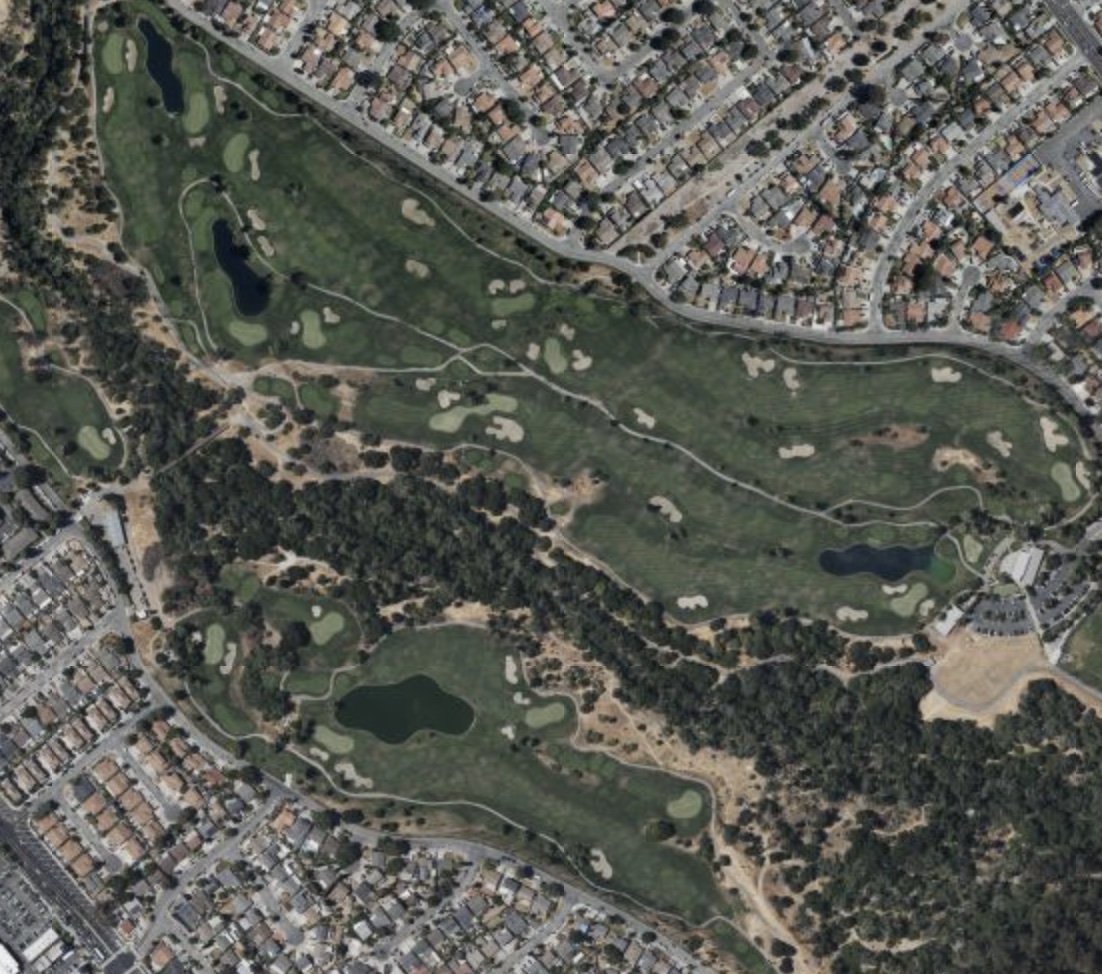}
        \includegraphics[width=0.225\textwidth]{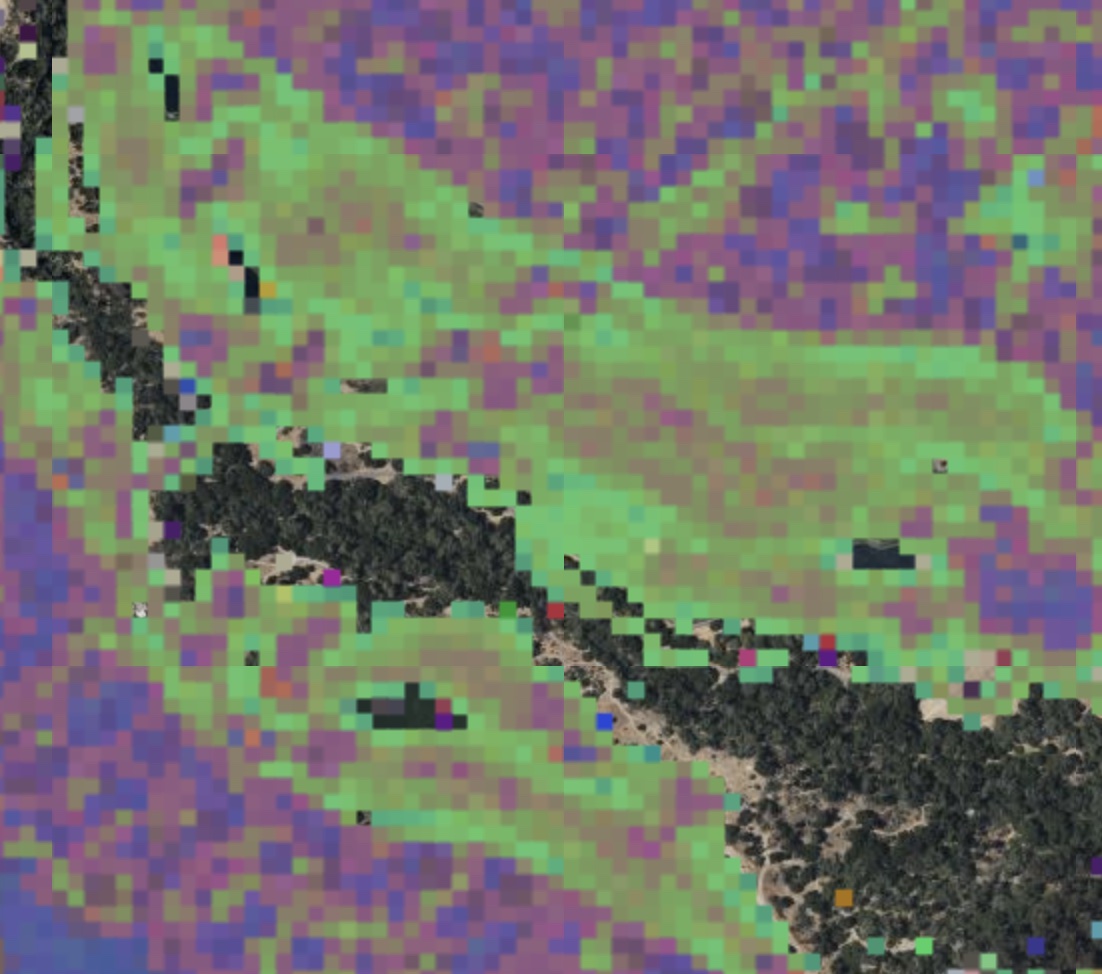}
        \centerline{Golf courses}
        \label{figure:gradcam_03}
    \end{minipage}

    \begin{minipage}{0.48\textwidth}
        \centering
        \includegraphics[width=0.225\textwidth]{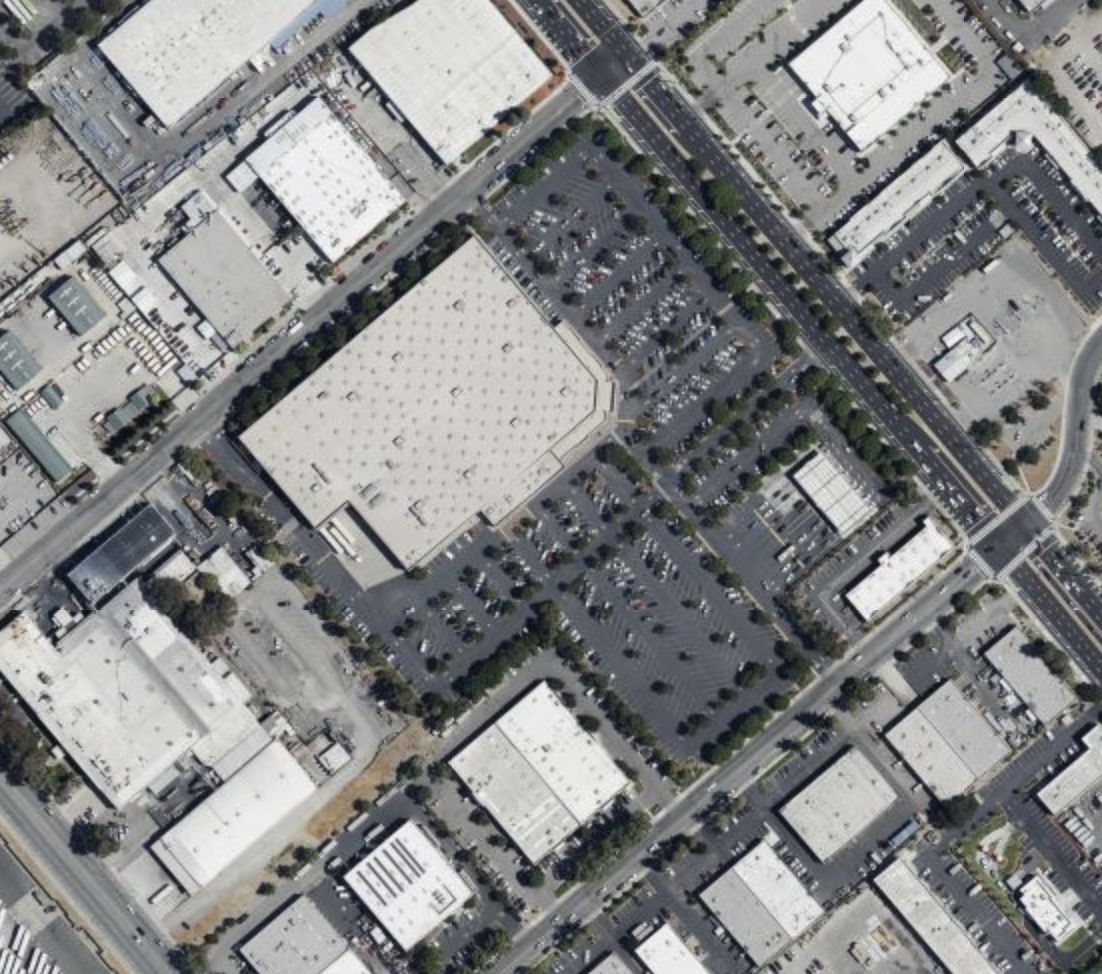}
        \includegraphics[width=0.225\textwidth]{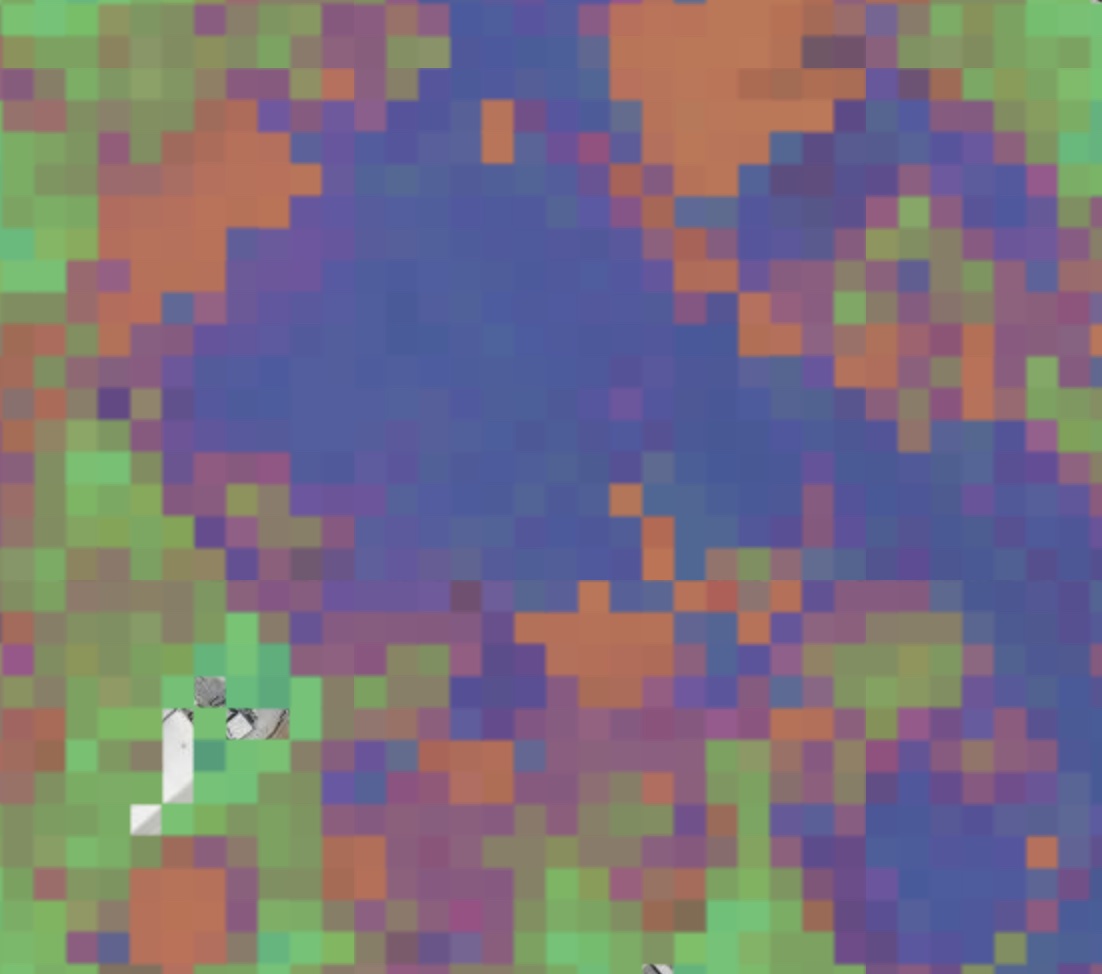}
        \includegraphics[width=0.225\textwidth]{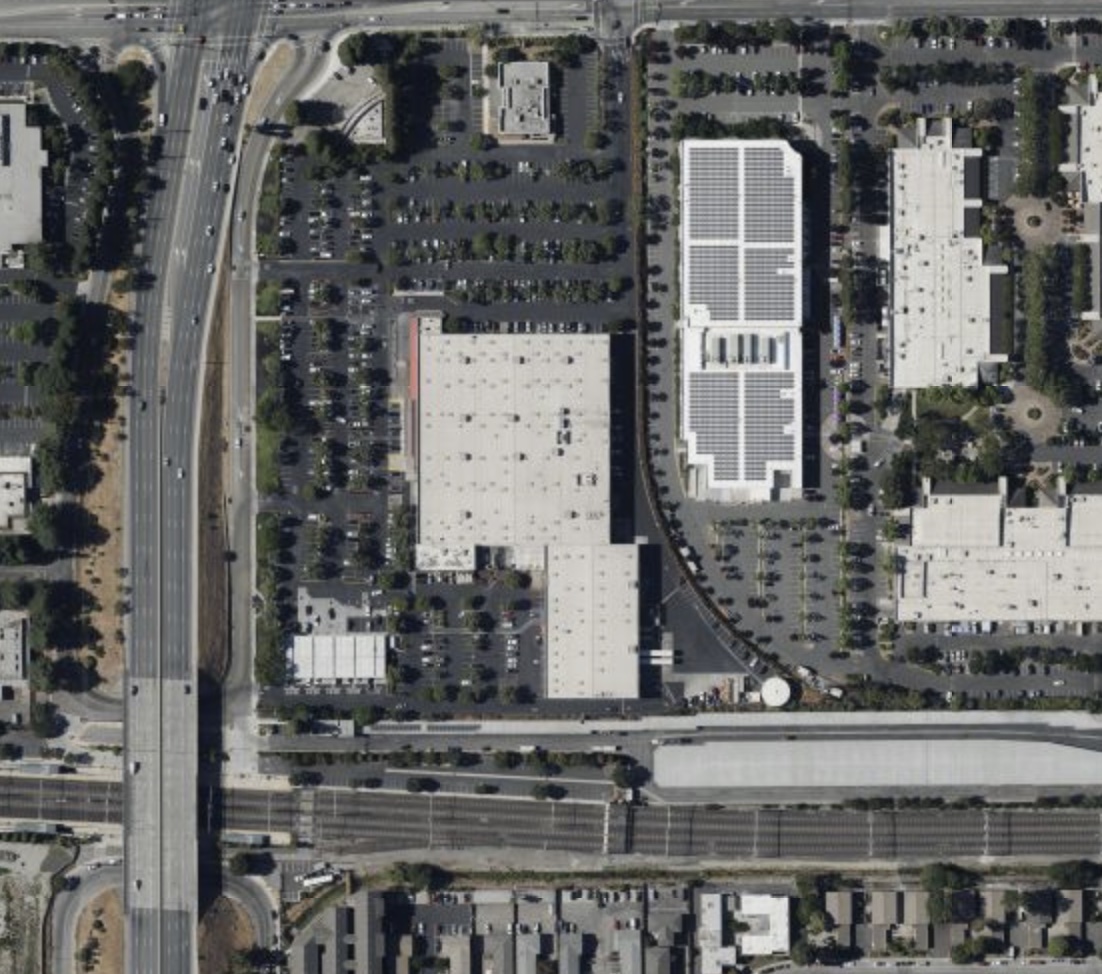}
        \includegraphics[width=0.225\textwidth]{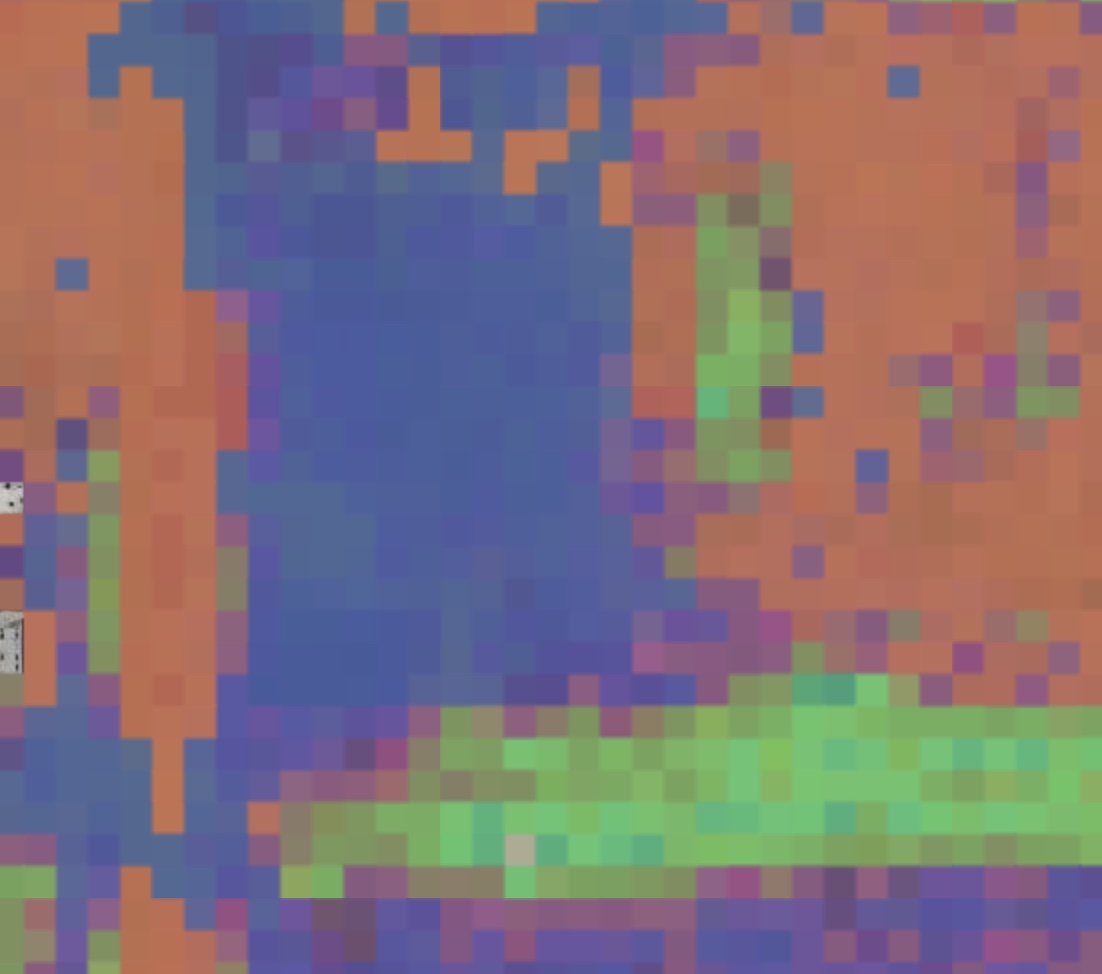}
        \centerline{Grocery shops}
        \label{figure:gradcam_04}
    \end{minipage}
    
    \begin{minipage}{0.48\textwidth}
        \centering
        \includegraphics[width=0.225\textwidth]{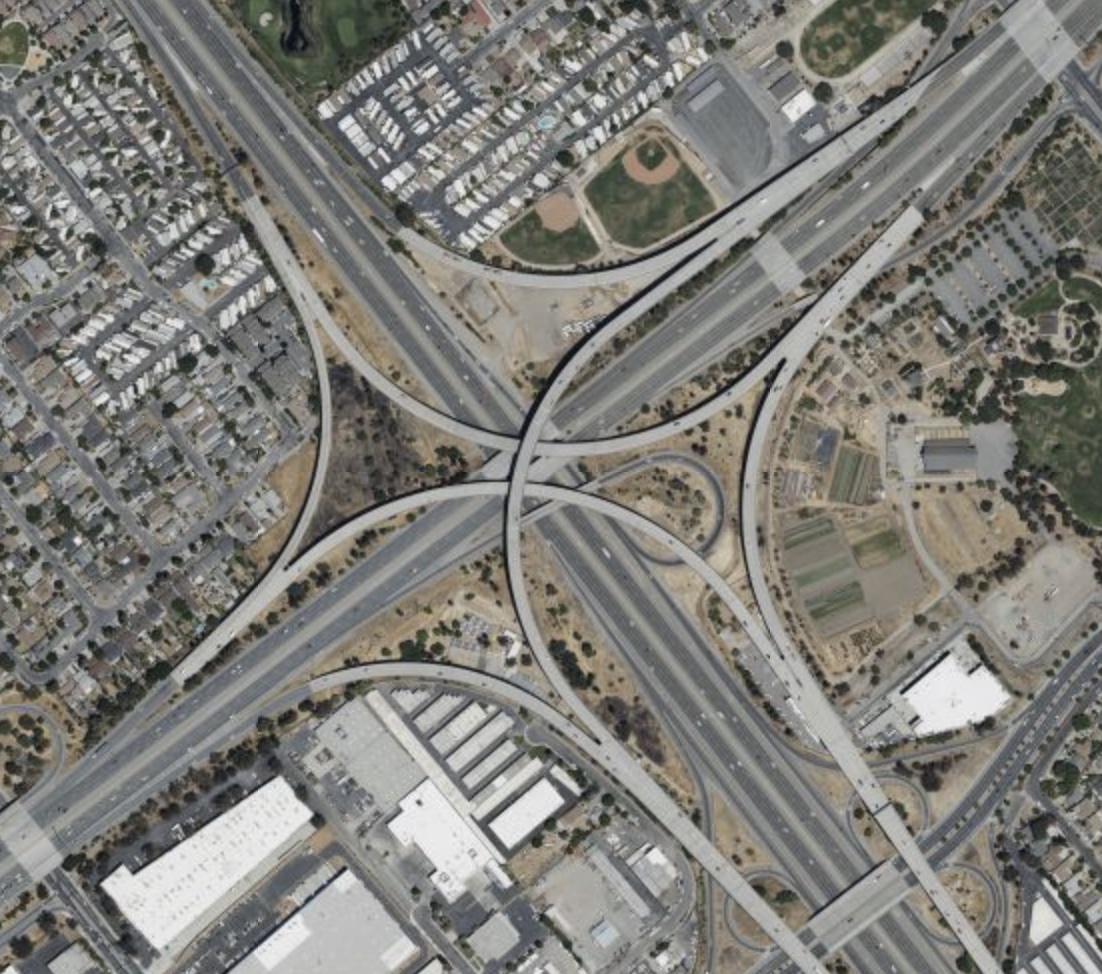}
        \includegraphics[width=0.225\textwidth]{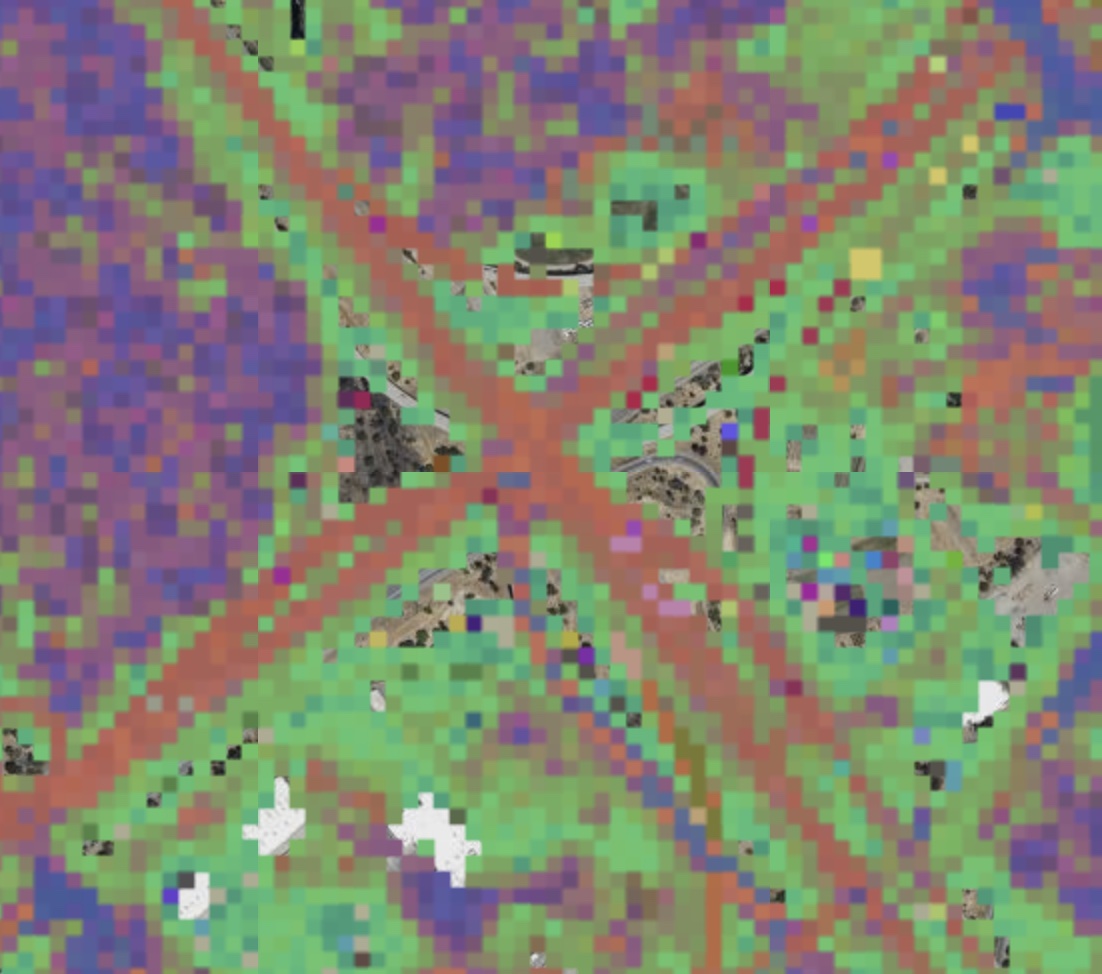}
        \includegraphics[width=0.225\textwidth]{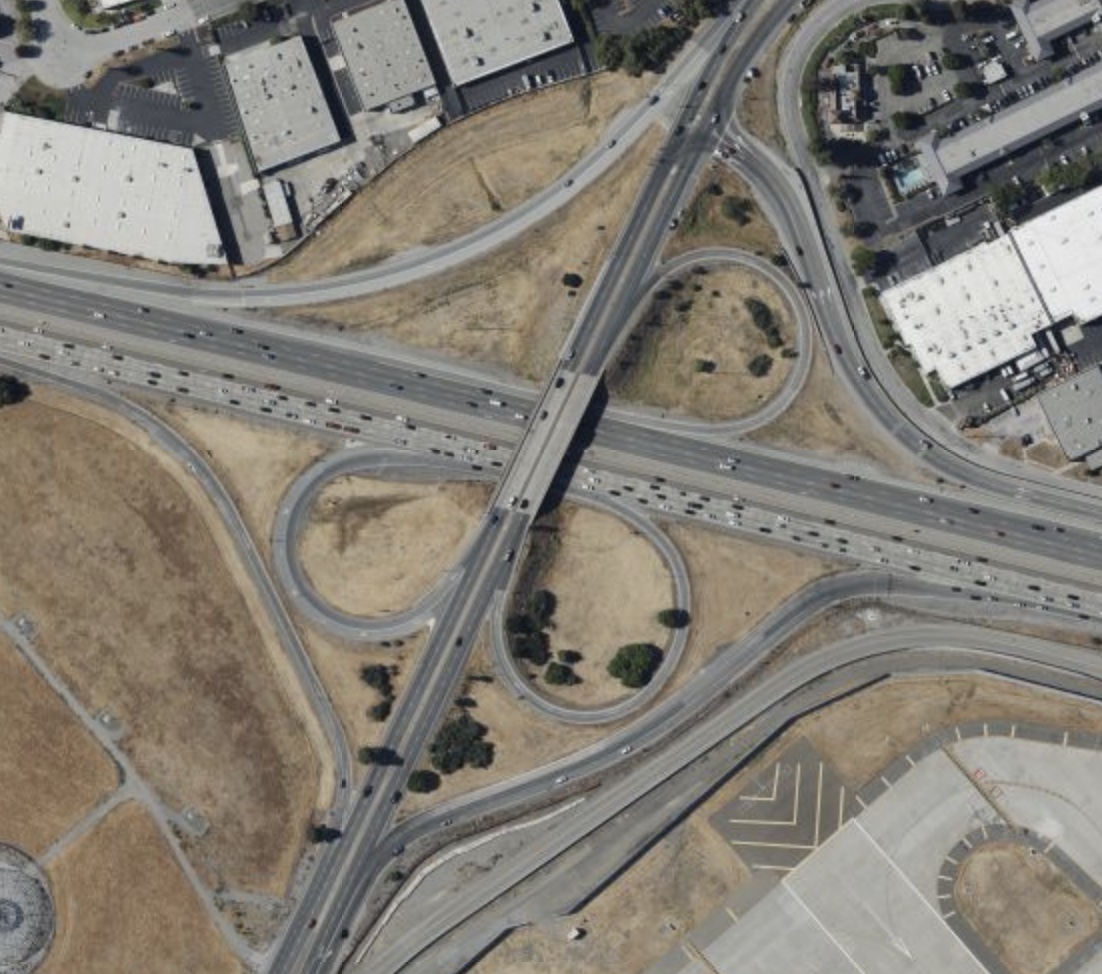}
        \includegraphics[width=0.225\textwidth]{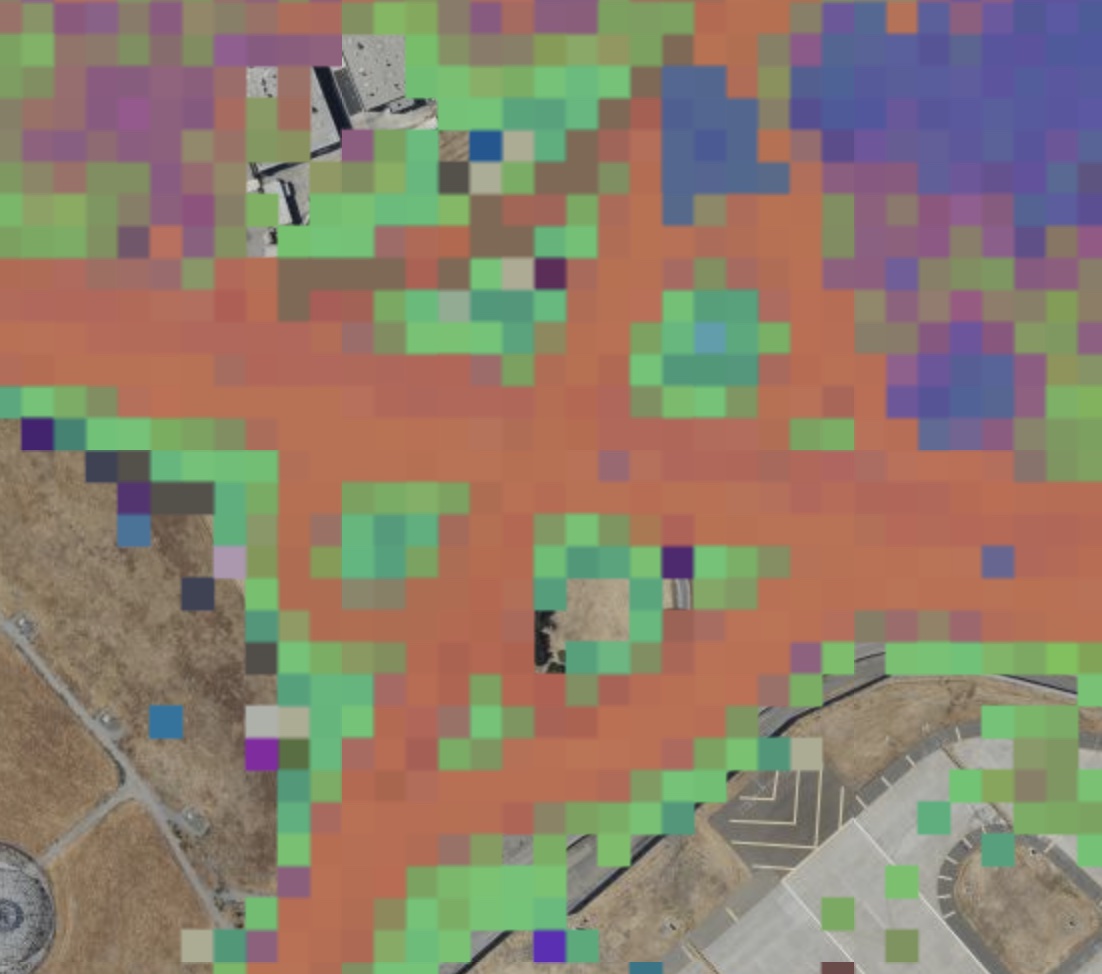}
        \centerline{Road Intersections}
        \label{figure:gradcam_05}
    \end{minipage}
    
    
    \caption{Two examples each of similar places having similar temporal patterns.}
    \label{figure:examples_of_stratification}
\end{figure}

To verify that the proposed embeddings do encode temporal patterns, we project the embeddings using UMAP to 3-dimensions and color the embedding space using RGB colors such that we can visualize the embeddings. Figure \ref{temporal_similarity}(a) shows the embeddings space, where each data point corresponds to a z24 tile and is color-coded per its position in the RGB space. Figure \ref{temporal_similarity}(b) is a geospatial map view of the z24 tiles colored by the same color scale as Figure \ref{temporal_similarity}(a). The clusters of color indicates that the embeddings are able to differentiate land uses. Figure \ref{figure:examples_of_stratification} shows a two examples each of zoomed-in views of different geographic locations. It is clear that locations wth similar land use share similar temporal patterns, which are encoded in the learned temporal embeddings.

\setlength{\belowcaptionskip}{2pt}
\setlength{\textfloatsep}{2pt}

\section{Applications} \label{applications}
In maps, residential area and commercial area are important contextual layers for task prioritization. Previously, we shows that commercial area generally observes heartbeat-like temporal patterns. This observation leads to a residential/commercial classification. The goal is to identify clusters of residential buildings and commercial places. We employ a UNet \cite{ronneberger2015u} within an in-house deep learning platform called the Trinity \cite{trinity, iyerperspectives} to conduct segmentation on zoom 16 tiles (\(384\times384\) meter) with zoom 24 tiles as pixels. The input is temporal embeddings of zoom 24 tiles with 32 channels ($d_r = 32$), and the output is a map of binary classes.

\begin{figure}[h]
    \centering
    \includegraphics[width=0.28\textwidth]{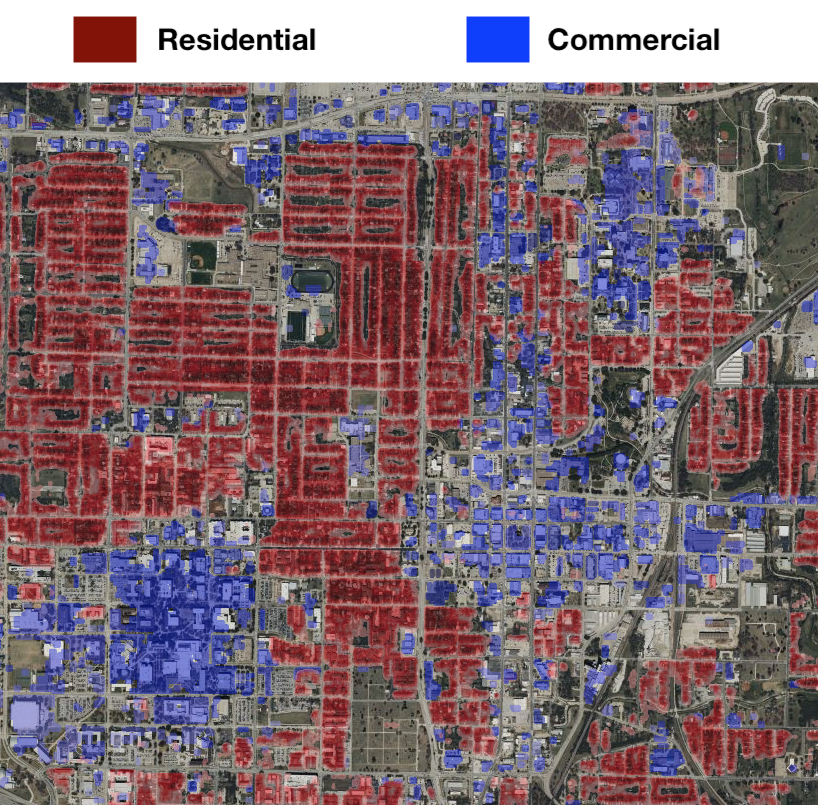}
    \caption{Residential and Commercial segmentation.}
    \label{residential}
\end{figure}

\begin{figure}[h]
    \centering
    \includegraphics[width=0.28\textwidth]{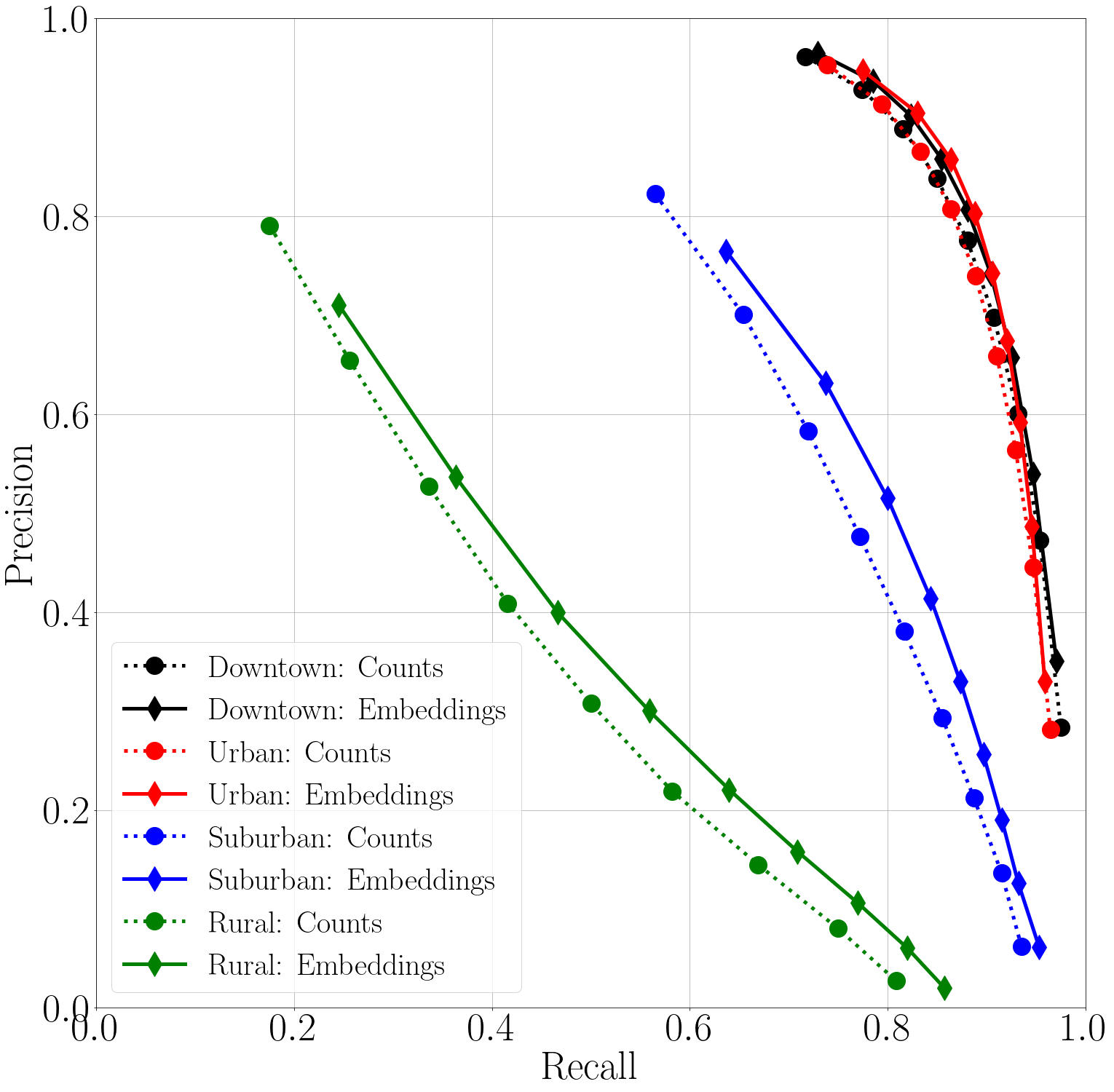}
    \caption{Performance comparison of the count with respect to temporal embeddings.}
    \label{metrics}
\end{figure}

Labels are manually collected. Polygons are drawn based on satellite image to cover residential blocks. Given that the concepts of residential and commercial are high-level and broad, we define residential area as clusters of private properties that are mainly built for living, knowing there are ambiguous cases like apartment rental or senior centers, which are residential in terms of functionality but commercial in terms of proprietorship. We also do not distinguish mixed-use structures where street fronts are commercial complex and high rising parts are residential apartments. Identifying mixed-use buildings required editors to scrutinize buildings one by one based on street view images which is costly. More practically, the labeling is solely based on satellite images and editors judge solely based on rooftops. We estimate there are roughly less than \(7\%\) of data errors in the ground truth set. However, ML models have the capability to tolerate a fractions of imperfect labels while generalizing the major patterns.

The classifier is set to run in multi-class mode, each pixel (zoom 24) is classified as one of three classes with a numeric confidence: residential, commercial, or background meaning no desired class is detected due to lack of confidence. Figure \ref{residential} shows residential and commercial predictions with a 0.5 confidence threshold. The blue footprint is a commercial complex surrounding by residential blocks in red. It is clear that the predictions are sufficient to depict reasonable boundaries on "area" level.

To quantify model performance, we calculate area overlap between tiles and ground truth data. Figure \ref{metrics} shows precision-recall curves of the model in different urban landscapes. To build a benchmark, we trained another model using simple aggregate counts as input. The embedding-based model outperforms the count-based model in all landscapes which is expected, because DFT catches the dynamics of activity in addition to the volume of activity, therefore the embedding-based model has more informational input than the count-based model. The difference of performance diminishes in downtown and urban environment because residential and commercial buildings tends to mix together, as a result, the temporal patterns are not as clear-cut as in suburb and countryside.

\section{Conclusion}\label{conclusion}
In this paper, we analyze time series extracted from mobility data and their temporal patterns. We show that using DFT for temporal analysis is more effective than count-based approaches. We also introduce an embedding framework that enables computer vision model to process time series data and accomplish geographic segmentation.

\bibliographystyle{IEEE}
{\footnotesize\bibliography{Paper_ACMSIGSPATIAL2021}}

\end{document}